%% file: main.tex
\documentclass{article}

\PassOptionsToPackage{%
  colorlinks=true,%
  pdfborder={0 0 0}
}{hyperref}

\PassOptionsToPackage{numbers, compress}{natbib}

  \usepackage[dblblindworkshop, final]{neurips_2025}

\usepackage{placeins}
\newcommand{\ourtitle}{The Amazing Stability of Flow Matching}
\usepackage{graphicx}

\usepackage{caption}
\usepackage{subcaption}
\usepackage{float} 
\usepackage{flafter} 
\input{preamble}

\title{\ourtitle}
\workshoptitle{}

\author{%
	Rania Briq$^{1,2}$ \\
	\texttt{r.briq@fz-juelich.de}
	\And
	Michael Kamp$^{2,3,4}$\\
	\And
	Ohad Fried$^{5}$\\
	\And
	Sarel Cohen$^{5}$\\
	\And
	Stefan Kesselheim$^{1,6,7}$\\
	\AND
	$^{1}$Forschungszentrum Jülich\\
	$^{2}$TU Dortmund\\
	$^{3}$Lamarr Institute\\
	$^{4}$Institute for AI in Medicine, UK Essen\\
		$^{5}$Reichman University\\
			$^{6}$Helmholtz AI\\
						$^{7}$University of Cologne\\
}

\begin{document}

\maketitle

\begin{abstract}
    The success of deep generative models in generating high-quality and diverse samples is often attributed to particular architectures and large training datasets. In this paper, we investigate the impact of these factors on the quality and diversity of samples generated by \emph{flow-matching} models. Surprisingly, in our experiments on CelebA-HQ dataset, flow matching remains stable even when pruning 50\% of the dataset. That is, the quality and diversity of generated samples are preserved. Moreover, pruning impacts the latent representation only slightly, that is, samples generated by models trained on the full and pruned dataset  map to visually similar outputs for a given seed. We observe similar stability when changing the architecture or training configuration, such that the latent representation is maintained under these changes as well.
    Our results quantify just how strong this stability can be in practice, and help explain the reliability of flow-matching models under various perturbations.

\end{abstract}

\input{sec_introduction}
\input{sec_method}
\input{sec_experiments}
\input{sec_discussion}

\clearpage

\FloatBarrier 
\clearpage
\bibliographystyle{plainnat}
\bibliography{references}

\end{document}

%% file: preamble.tex
\usepackage{amsmath, amsthm, amssymb}
\usepackage{natbib}

\usepackage[utf8]{inputenc} 
\usepackage[T1]{fontenc}    
\usepackage{hyperref}       
\usepackage{url}            
\usepackage{booktabs}       
\usepackage{amsfonts}       
\usepackage{nicefrac}       
\usepackage{microtype}      

\usepackage{xcolor}
\definecolor{darkblue}{rgb}{0.0,0,0.75} 
\usepackage{wrapfig}

\hypersetup{ 
  urlcolor = darkblue, 
  linkcolor = darkblue, 
  citecolor = darkblue, 
  pdftitle=\ourtitle, 
  breaklinks 
}

\newcommand{\paragraphit}[1]{\noindent\textbf{\textit{#1}}\;}


%% file: sec_introduction.tex
\section{Introduction}
\label{sec:intro}
Diffusion models \citep{song2019generative, song2020score} have driven tremendous advances in generative modeling over the past few years~\cite{rombach2022high,ho2022imagen,blattmann2023align,liu2402sora}. Flow-matching (FM) methods, an alternative to diffusion models \cite{ho2020denoising}, promise several advantages in efficiency and simplicity while achieving competitive performance \cite{lipman2022flow,liu2022flow,tong2023improving,bar2024lumiere,liu2402sora}. Yet, training these models remains demanding both in compute and data, therefore, tailoring a dataset to desired generative properties has the potential to significantly reduce the computational cost. In this paper, we study the stability of FM under data perturbation, and develop informed approaches to data pruning. Using standard metrics, we probe stability and find a surprising result: even under strong perturbation of data and model architecture, trajectories initialized with the same random noise evolve to visually similar samples.

For diffusion models, stability in generating high-dimensional realistic data has been observed in several works: \citet{kadkhodaie2024generalization} observed that diffusion models produce similar outputs under the same seed when trained on two disjoint subsets of the data, arguing that different splits converge to a similar geometry-aligned basis that follows image contours. \citet{mlodozeniec2024influence} confirmed the stability phenomenon for diffusion models while studying data attribution, and report that the likelihood remains nearly constant across models trained on 50\% random subsets. However, this was not shown for FM models, nor was it shown how this invariance can be exploited to train diffusion models on less data.
Furthermore, in diffusion models, the objective is related to entropic-transport optimization (Schrödinger bridge) \citep{de2021diffusion}, which is stable to data perturbations \citep{ghosal2022stability}. By contrast, the flow-matching objective fits a velocity field $u(x,t)$, whose ODE transports noisy latents towards a clean manifold. While diffusion's stability is theoretically grounded, the behavior of FM models under data and architectural perturbations remains largely underexplored.

We empirically analyze the behavior of FM models trained on subsets of data and with different architectures.

We summarize our contributions as follows:
\vspace{-\parskip}
\begin{itemize}
  \setlength{\topsep}{0pt}
  \setlength{\itemsep}{0pt}
  \setlength{\parsep}{0pt}
  \setlength{\partopsep}{0pt}

    \item We show that FM models are remarkably stable. The generated images are visually similar under a wide variety of perturbations, including training on disjoint subsets of the dataset, swapping of the entire dataset, as well as model architecture shrinkage. Each perturbation affects the generated data semantically only minimally.
    \item Inspired by previous work in discriminative models, we introduce three informed data pruning methods to FM models and study their influence qualitatively and quantitatively. 
    \item Our proposed cluster-based resampling method that balances the distribution between different clusters can even improve the evaluation metrics of the generated images. 
\end{itemize}
\vspace{-\parskip}

%% file: sec_method.tex
\section{Approach}
We probe the stability of FM models under various perturbations to the training data distribution, as well as the model architecture. Data perturbations include dataset pruning, where given a dataset $S$, we find a subset $S'\subset S$  using the pruning methods proposed in \cite{briq2024data}. We (i) use a \emph{random} subset as a baseline, whose performance serves as a lower bound for methods that require computation; (ii) rank based on a sample's training signal: gradient norm or loss computed along shared noise paths and timesteps; and (iii) cluster samples using their semantic features in a pretrained embedding space. For each method, we also apply the inverse criterion, i.e. we select samples with the lowest scores instead of the highest ones, and denote it by the superscript $-1$.
\vspace{-0.6\baselineskip}
\subsection*{Pruning methods}\label{sec:pruning-methods}
\vspace{-0.6\baselineskip} 
\paragraphit{Gradient-based scoring (Grad)}. 
Under this strategy, we train a small surrogate model $\approx 7\%$ of the full training schedule, and use it to estimate the gradient magnitude for each sample using $M=2$ fixed random noisy samples and $T=8$ timesteps, creating shared noise paths for all the samples and decreasing the variance stemming from randomness. The gradient norms are then averaged over $M$ and $T$ using exponential moving average (EMA) estimate per $t\in T$ to remove large scale bias of larger noise bands.
\begingroup
\setlength{\abovedisplayskip}{6pt}
\setlength{\belowdisplayskip}{6pt}
\setlength{\abovedisplayshortskip}{0pt}
\setlength{\belowdisplayshortskip}{3pt}

\begin{equation}
	s_i^{\mathrm{grad}}
	= \frac{1}{T}\sum_{k=1}^{T}\frac{1}{M}\sum_{m=1}^{M}
	\frac{\left\|\nabla_\theta \,\ell\!\big(x_i;\, t_k, x_0^{(m)}\big)\right\|_2^2}{\mu_g(t_k)}\,,
	\label{eq:s_grad}
\end{equation}
where $\mu_g(t_k)$ is the EMA estimate at timestep $t_k$ over samples and noise endpoints of the squared gradient norm $\|\nabla_\theta \ell(x_i; t_k, x_0^{(m)})\|_2^2$ with respect to the model parameters $\theta$, computed inside the loop, and $x_0^{(m)}$ are the shared noise endpoints.
 The computation is easily parallelizable, however, this is an expensive method and we only apply it to gain insights into the effect of high-gradient samples on the model. Since samples with a large gradient influence the learned velocity field, we expect retaining them has a positive impact on the model.
 
\vspace{-0.2\baselineskip} 
\paragraphit{Loss-based scoring (Loss).}
We apply the same setup used in \emph{Grad} and define $s_i^{\text{Loss}}$ similarly,  replacing $\|\nabla_\theta \ell\|_2^2$ by $\ell$ and $\mu_g$ by $\mu_\ell$.

\vspace{-0.2\baselineskip} 
\paragraphit{Cluster-based scoring (Clust).}
We extract the image embeddings using the pretrained visual model CLIP \cite{radford2021learning}). We then use k-means~\citep{lloyd1982least} to cluster the samples, producing groups that share similar semantic characteristics. There are two criteria to consider here: (i) how many samples to select from a cluster, and (ii) which samples.  For (i), we select either a number \emph{proportional} to the cluster size or a \emph{balanced} number, i.e. selecting an equal number of samples from each cluster. The first inherits the underlying distribution imbalance, while the latter balances skewed datasets.
For (ii), we score a cluster's population based on their distance from the cluster center, and select either those located nearest to its center or furthest. The nearest samples form a representative subset retaining the core characteristics of the distribution, while the furthest samples cover more difficult and scarce samples. We refer to these variants as \textit {Clust$_{p/b}^{1/-1}$}, indicating nearest/furthest and proportional/balanced. In our experiments, we choose $k=24$ based on analysis of the clusters' inertia. 

%% file: sec_experiments.tex
\section{Experiments}

\begin{figure}[!h]
	\centering
	\captionsetup[sub]{skip=2pt}
	
	\begin{minipage}[t]{0.475\columnwidth}
		\centering
		\begin{subfigure}[t]{\linewidth}
			\centering
			\includegraphics[width=\linewidth]{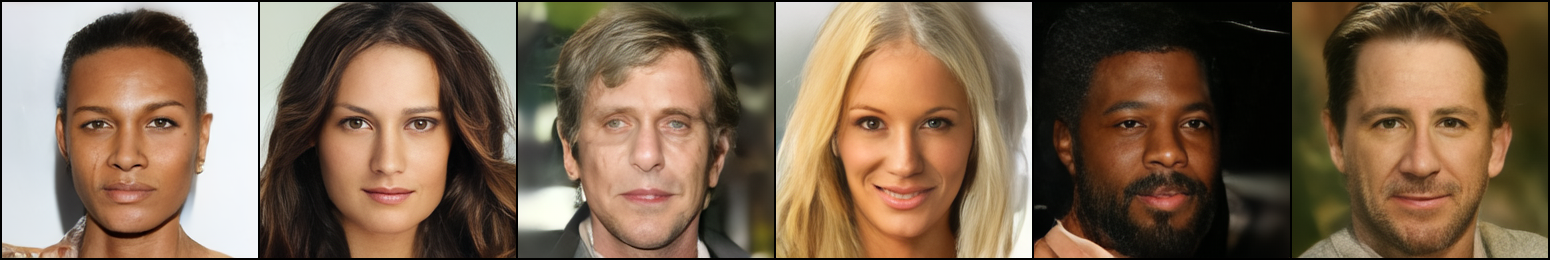}
			\includegraphics[width=\linewidth]{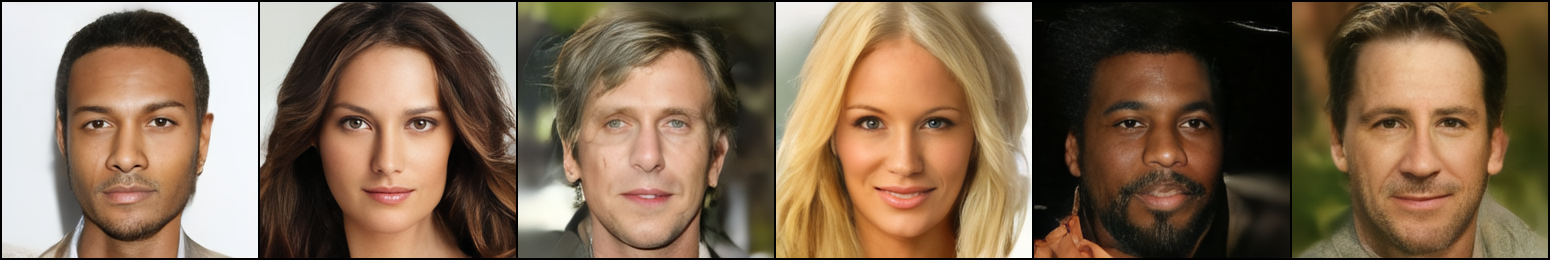}
			\caption{Two disjoint random subsets}
			\label{fig:complementary_data}
		\end{subfigure}
		
		\vspace{2mm} 
		
		\begin{subfigure}[t]{\linewidth}
			\centering
			\includegraphics[width=\linewidth]{images/set1/vanilla_00003_epoch80k}
			\includegraphics[width=\linewidth]{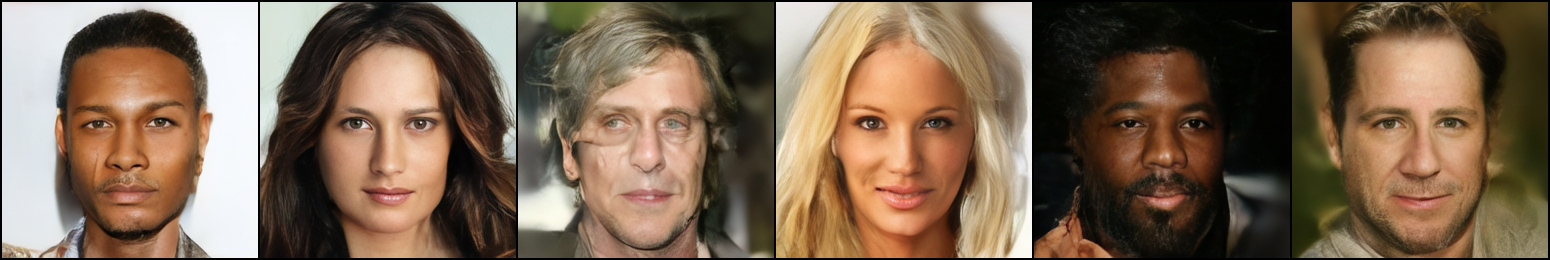}
			\includegraphics[width=\linewidth]{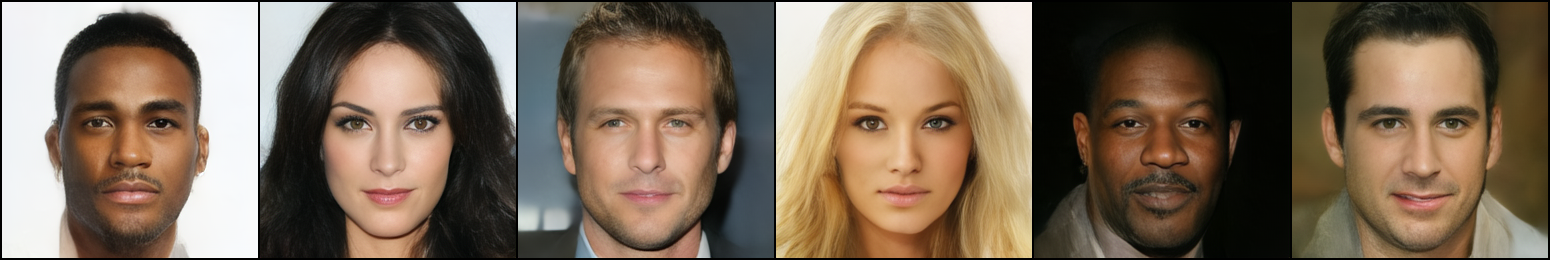}
			\caption{DiT-XL/4 $\rightarrow$ DiT-S/2 $\rightarrow$ U-Net}
			\label{fig:arch_xl2_s4_unet}
		\end{subfigure}
	\end{minipage}
	\hfill
	\begin{minipage}[t]{0.475\columnwidth}
		\centering
		\begin{subfigure}[t]{\linewidth}
			\centering
			\includegraphics[width=\linewidth]{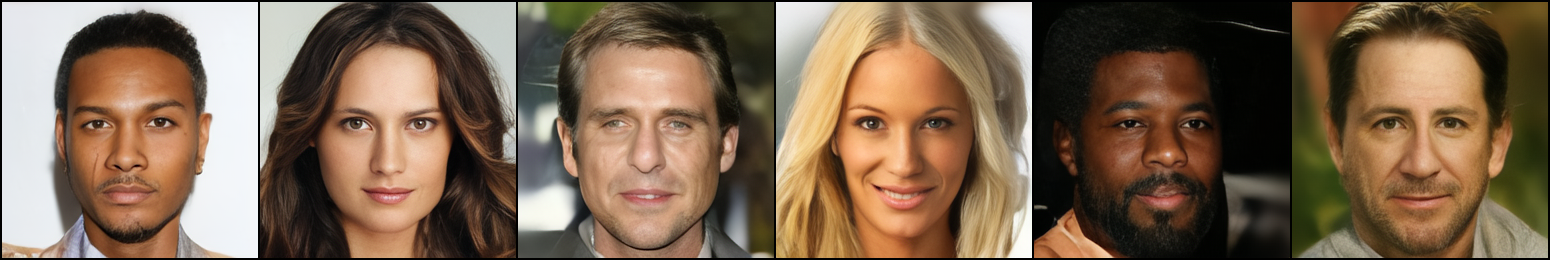}
			\includegraphics[width=\linewidth]{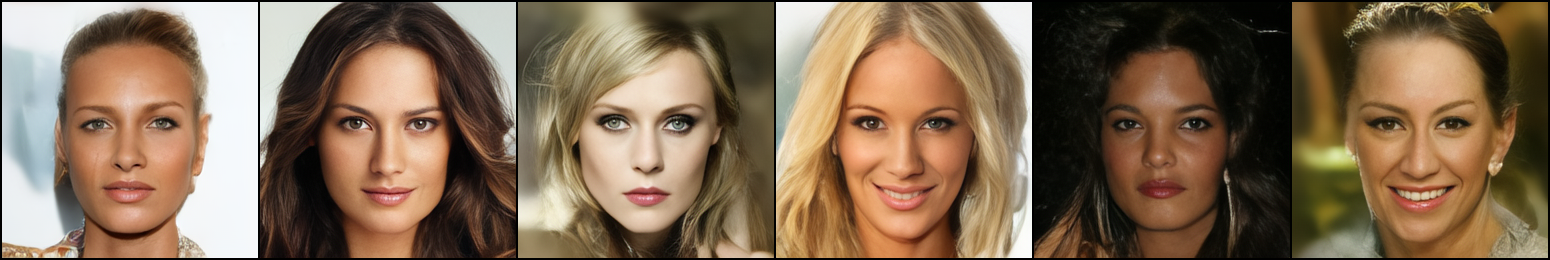}
			\includegraphics[width=\linewidth]{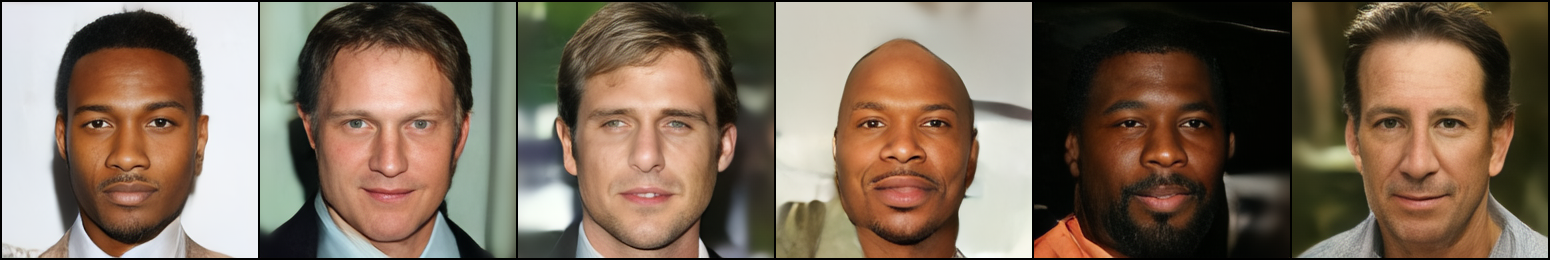}
			\caption{Both genders/Female/Male}
			\label{fig:female_male_mixed}
		\end{subfigure}
		
		\vspace{2mm} 
		
		\begin{subfigure}[t]{\linewidth}
			\centering
			\includegraphics[width=\linewidth]{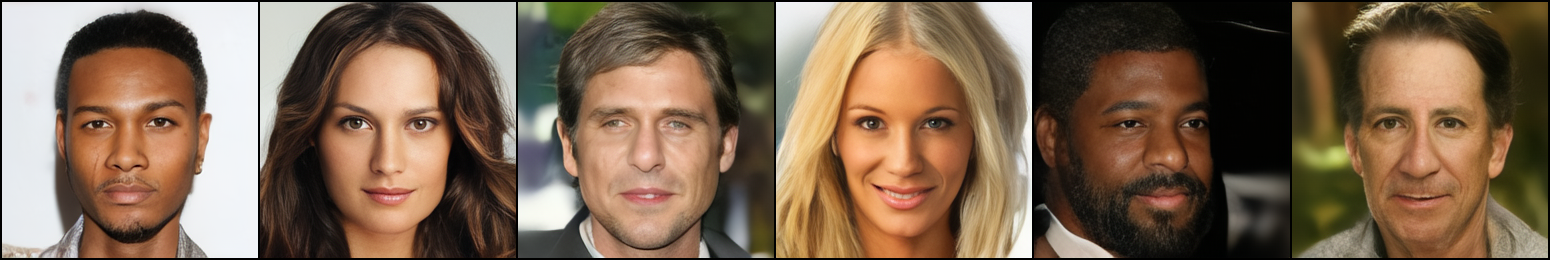}
			\includegraphics[width=\linewidth]{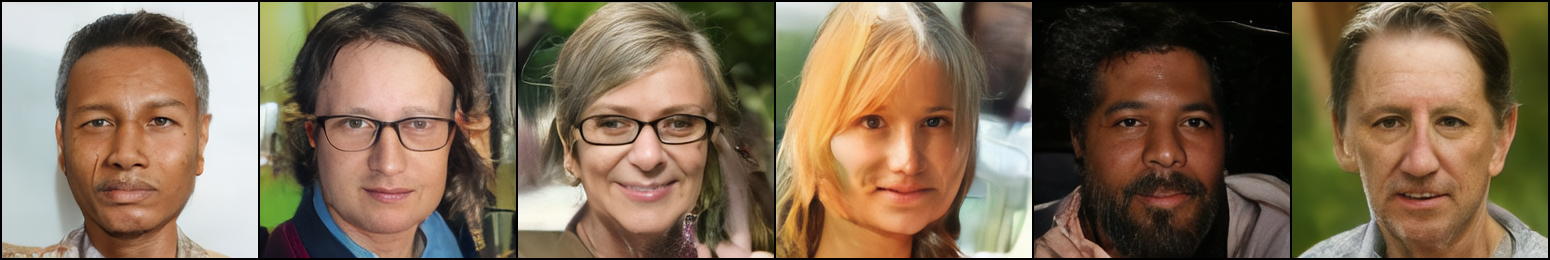}
			\caption{CelebHQ $\rightarrow$ FFHQ}
			\label{fig:fmffhq_vaeceleb}
		\end{subfigure}
	\end{minipage}
	
	\caption{Stability of the generated images. (a) We train the model on two disjoint random subsets of the data, and obtain visually very similar images.
 (b) Model capacity change from DiT-XL to DiT-S retains high similarity, while switching to a U-Net architecture retains similarity to a lesser degree. (c) The data is split into two sets based on zero-shot binary classification of gender (male/female): Images we visually interpret as belonging to the retained partition are semantically preserved, while images of the complementary class swap the semantic interpretation. (d) Changing the training dataset from CelebHQ to FFHQ, while still using CelebHQ VAE, retains similarity too.}
	\label{fig:stability_across_board}
\end{figure}

\begin{table}[hb]
	\centering
	\scriptsize
	\setlength{\tabcolsep}{3pt}
	\renewcommand{\arraystretch}{1.1}
	
	\begin{subtable}{\columnwidth}
		\centering
		\begin{tabular}{lccccccccccc}
			\toprule
			& Unpruned & Random & Grad & Grad$^{-1}$ & Loss & Loss$^{-1}$ & Clust$_p$ & Clust$_p^{-1}$ & Clust$_b$ & Clust$_b^{-1}$ \\
			\midrule
			FID $\downarrow$ &
			24.24 & 25.25$\pm$0.38 & 24.62 & 29.75 & 33.92 & 23.49 & 25.19 & 27.96 & \textbf{22.80} & 26.77 \\
			\bottomrule
		\end{tabular}
		\vspace{2mm}
		\caption{FID on CelebHQ with 4k generated samples at $pr=0.5$. Random is averaged over 3 seeds.}
		\label{tab:allmethods_pr05_metrics}
	\end{subtable}
	
	\vspace{2mm}
	
	\begin{subtable}{\columnwidth}
		\centering

		\begin{tabular}{lcccccccccc}
	\toprule
	Random & Grad & Grad$^{-1}$ & Loss & Loss$^{-1}$ & Clust$_p$ & Clust$_p^{-1}$ & Clust$_b$ & Clust$_b^{-1}$ \\
	\midrule
	$0.83\pm0.11$ &
	$0.79\pm0.12$ & 
	$0.80\pm0.12$ &
	$0.80\pm0.12$  & 
	$0.80\pm0.13$  &
	$0.80\pm0.12$ & 
	$0.81\pm0.12$  &
	0.81$\pm$0.12
	& $0.79\pm0.13$ 
		\end{tabular}
		\vspace{2mm}
	\caption{ArcFace cosine similarity between each pruned model and \emph{Unpruned} for $pr=0.5$. $N=4k$ pairs matched by seed are evaluated. Here, $\pm$ denotes standard deviation over image pairs. Unmatched pairs yield $0.37\pm 0.11$.
	}
	\label{tab:allmethods_pr05_arcface_simlarity}
	\end{subtable}
	
\end{table}

\subsection*{Experimental setup}
\vspace{-0.6\baselineskip} 
We use the transformer-based architecture DiT \cite{peebles2023scalable}, and replace diffusion with flow-matching transport \cite{esser2024scaling} (we name it FM-DiT), training a velocity field $u_\theta(x,t)$ along linear interpolants between Gaussian noise and the data. We also train a vector-quantized variational autoencoder (VQ-VAE) \cite{van2017neural} using the same target dataset to encode the images, similar to Stable Diffusion \cite{rombach2022high}\footnote{Code and experimental details are available at~\url{https://github.com/briqr/fm_stability}.}.
DiT is based on a ViT-style transformer \citep{dosovitskiy2020image}, which operates on image patches with global self-attention. For the architectural change experiment, we additionally train a U-Net backbone \citep{ronneberger2015u}, following multi-scale convolutional encoder–decoder with skip connections as done in diffusion models \citep{ho2020denoising,rombach2022high}.

For quantitative evaluation, we report FID \cite{heusel2017gans}, which measures the Fréchet distance between feature embeddings of the generated and training distributions. For quantifying FM stability, we measure ArcFace pairwise cosine similarity for faces \cite{deng2018arcface}, a standard embedding model for face identification. \emph{Unpruned} refers to the model trained on the full dataset. 
 $N=4096$ denotes the number of generated samples. All experiments are carried out by training the respective models on the standard CelebHQ dataset~\citep{karras2017progressive}, which is based on CelebA dataset \citep{liu2015deep}. We acknowledge the imbalance in the dataset, which can affect qualitative judgments and subgroups performance.
We further emphasize that features and attributes of human faces are subjective. 
All reported images are chosen from the same sequence of random $x_0 \sim \mathcal{N}(0, I)$. The images were not hand-picked; we only selected a range inside a longer sequence.

\subsection*{Stability tests}

We investigate FM stability using several stress tests, including substantial data perturbation through pruning and data swapping, and architectural changes either in the model capacity or architecture design.

\vspace{-0.2\baselineskip} 
\paragraphit{Disjoint subsets.} In Fig.~\ref{fig:complementary_data}, we train two FM model instances on two disjoint random subsets of the data. When integrating the velocity field starting from the same random points $x_0$, we observe that the outputs are nearly identical. We quantify this consistency using ArcFace pairwise similarity, and obtain $sim=0.69\pm0.12$ between $N=4096$ sample pairs generated by both models, where $\pm$ denotes the standard deviation over pairs. For comparison, unrelated pairs yield $sim=0.34\pm0.10$. 

\paragraphit{Cluster removal.}Fig.~\ref{fig:female_male_mixed} depicts another experiment that alters the training data substantially. The first FM model is trained on images classified as female by PaliGemma VLM \citep{beyer2024paligemma}, while the second on images classified as male, yielding ArcFace similarity $0.76\pm 0.17$ and $0.58\pm 0.16$ respectively. This experiment is analogous to dropping an entire cluster or mode of the distribution. The results show that apart from the removed cluster, the models continue to generate similar outputs, demonstrating FM stability to mode removal. 
We want to acknowledge that we performed the binary split on the gender attribute as a technical experiment, and that the societal concept and implications of gender are clearly much more complex.

\begin{wrapfigure}{r}{0.475\textwidth}
	\begin{center}
		\includegraphics[width=\linewidth]{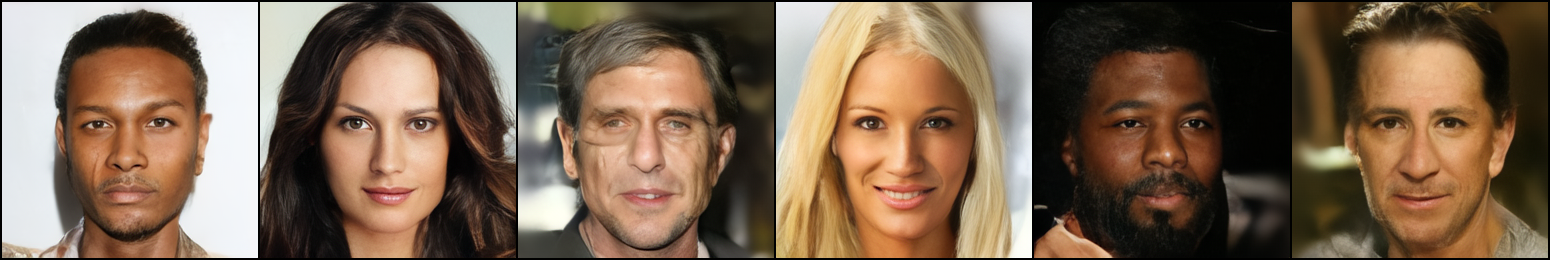}
		\includegraphics[width=\linewidth]{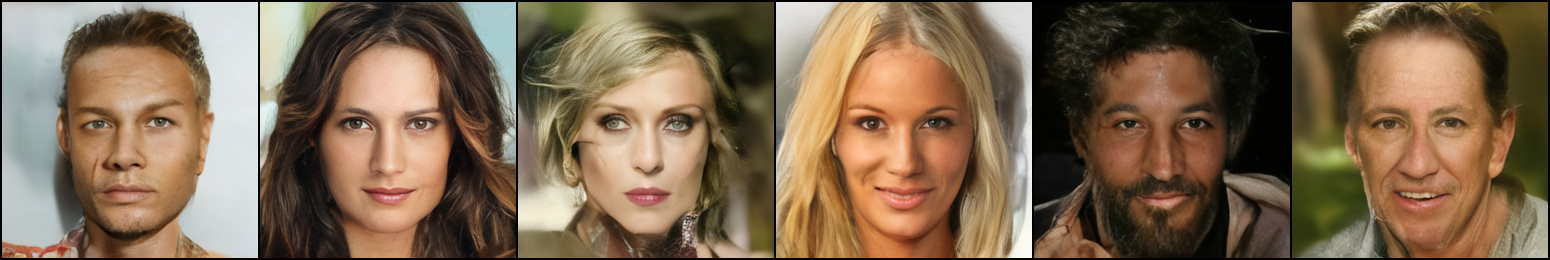} 
		
		\vspace{2mm}
		\includegraphics[width=\linewidth]{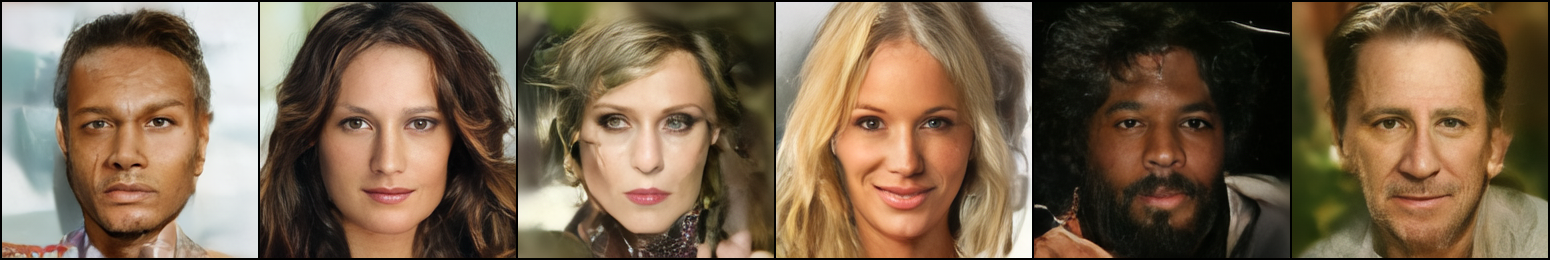}
		\includegraphics[width=\linewidth]{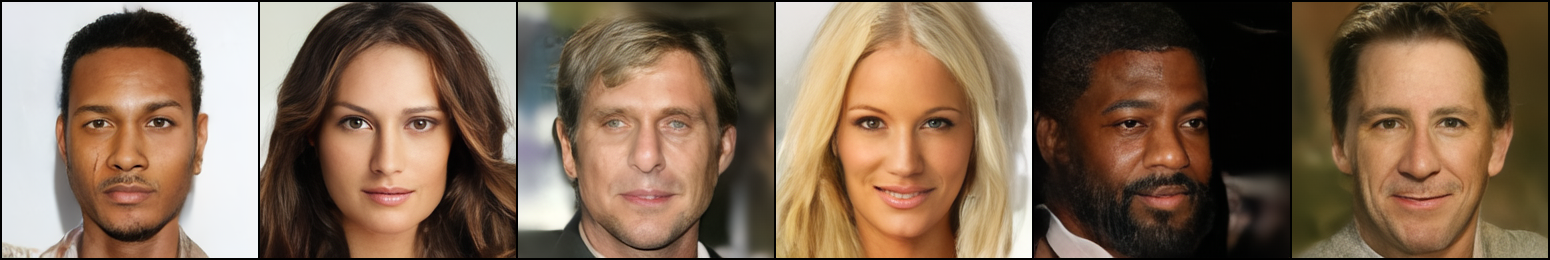} 
		
		\vspace{2mm}
		\includegraphics[width=\linewidth]{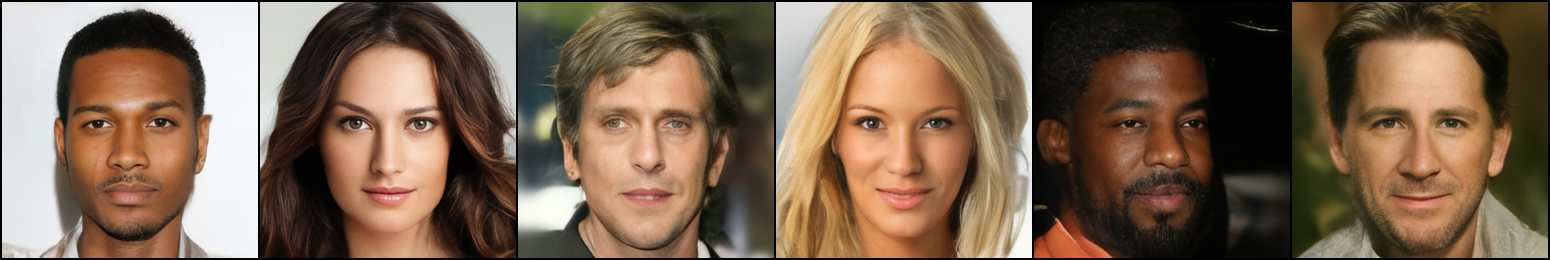}
		\includegraphics[width=\linewidth]{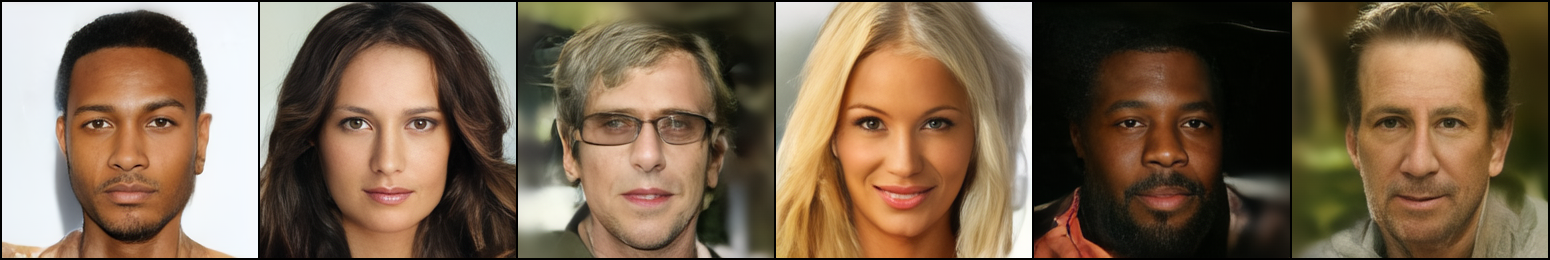}
	\end{center}
	\caption{Stability of the generated images under different pruning strategies and their inverse, $pr=0.5$.  In order: \emph{Grad$^{1/-1}$}, \emph{Loss$^{1/-1}$}, \emph{Clust$_b^{1/-1}$}.}
	\label{fig:all_pruning_methods_pr05}
\end{wrapfigure}

\paragraphit{Data swapping.} The experiment depicted in Fig.~\ref{fig:fmffhq_vaeceleb} is the most extreme form of data alteration. The FM model is trained on a different but same-domain dataset, FFHQ \citep{karras2019stylegan}, which also comprises human faces. Even then, the outputs retain resemblance and we obtain ArcFace similarity $sim=0.58\pm0.15$ (unrelated pairs yield $sim=0.30\pm0.10$), indicating that with a fixed latent space, a different but same-domain dataset such as FFHQ lies on the same manifold as CelebA-HQ, allowing FM models to continue to learn similar trajectories with a matching seed.

\paragraphit{Architectural change.} Fig.~\ref{fig:arch_xl2_s4_unet} illustrates a different type of stability tests. Instead of perturbing the distribution, we train three model variants that differ in their capacity or architecture. The first two variants share the same transformer DiT architecture but differ in their size: DiT/XL-2 (675M parameters, 24 layers) and  DiT/S-4 (33M, 12 layers). The third variant is based on U-Net architecture. The outputs were consistently similar under model capacity change, retaining ArcFace similarity $0.81\pm0.12$, indicating identity preservation. With the full architectural change, the similarity drops to $0.55\pm 0.13$. While the drop is clearly more visible compared to capacity shrinkage, we observe that the coarse attributes are preserved, which again hints at global stability.

\paragraphit{Pruning strategies.} In Fig.~\ref{fig:all_pruning_methods_pr05}, we apply the proposed pruning methods and their inverse. 
Even though \emph{Grad$^{-1}$} (row 2) and \emph{Loss} (row 3) produce artifacts, perceptually the images are similar to the inverse method.
We apply the pruning methods using pruning fraction $pr=0.5$ and report the results in table~\ref{tab:allmethods_pr05_metrics}. \emph{Random}'s FID deteriorates slightly from \emph{Unpruned}. Using \emph{Grad} almost does not change the FID, while its inverse (selecting lowest-grad samples) deteriorates significantly, as expected when dropping samples most influencing the model's weights. Selecting the highest-loss samples (\emph{Loss}) substantially worsens the FID, compared to discriminative models \cite{paul2021deep}. In FM, these samples' predicted velocity fields deviate from the target flow, which is typical of samples present in low-density regions. Increasing these samples' representation therefore would lead to adversely impacting the flow. This explains why \emph{Loss$^{-1}$} has the opposite effect. \emph{Clust$_b$} even improves the FID, thanks to its uniform coverage across clusters, indicating that performance does not only depend on the sheer amount of data, but also on how balanced the data is. Across pruned variants versus their inverse, we obtain ArcFace cosine similarity in the range $0.72-0.74\pm0.13$, compared to $0.37\pm 0.11$ for randomly shuffled pairs, indicating that matched outputs remain much closer than unmatched ones even when the training subsets are disjoint.\\
In table~\ref{tab:allmethods_pr05_arcface_simlarity}, we quantify the stability of FM by comparing each pruned variant with \emph{Unpruned}. We compare $N=4096$ pairs matched by seed and observe that all methods maintain high similarity (above 0.79), compared to unrelated pairs ($0.37\pm0.12$). This suggests FM models are very robust to perturbation in their training set: even methods that degraded the performance in FID, such as \emph{Loss} and \emph {Grad}$^{-1}$, maintained high similarity with  \emph{Unpruned}.

\label{sec:exp}

%% file: sec_discussion.tex
\section{Discussion and Outlook}
We interpret our observation in terms of how well our model learns to approximate the true velocity field. We observe that for models trained under various perturbations, when starting from the same initial point $x_0$, the trajectories obtained by integrating the flow ODE end in points $x_1$ that are very close, and decode to perceptually similar images.

Recent works have begun to investigate FM models' ability in generalization, for example, ~\citet{bertrand2025closed} show that learning using the derived closed-form of the velocity field \citep{gao2024flow} in the finite data regime yields a similar performance as when using stochastic target $u(x,t)$, suggesting that stochasticity is averaged out and is therefore not the source of generalization. Our experiments on stability are complementary to this view; despite extreme data perturbations and architectural changes, trajectories starting from the same noise converge to nearby endpoints, suggesting their generalization does not stem from a single factor.

We studied removing entire clusters within the data distribution. For this model, trajectories starting from $x_0$ that would have ended up in endpoints in this cluster for a model trained on the full dataset, were rerouted to different endpoints. The corresponding images are clearly different, while images from retained clusters remain similar.  In particular, trajectories sufficiently far away from the ones influenced by excluded clusters are only weakly impacted. 
We interpret this as a global stability: The flow field is only adjusted locally where necessary, while the global structure remains unaltered. This stability when removing data systematically allows enables training models with less data. 

We also show that dataset pruning can be performed with little negative impact, or even with positive impact when done correctly. Some methods exhibit strong adverse effects, which hints at an intricate interaction of the dataset choice and generalization of FM models. We believe that understanding this interplay better is of high relevance for future powerful generative models trained on very large amounts of data, and could improve their efficiency substantially.

\section*{Acknowledgments}
This work is funded by the German Federal Ministry
for Economic Affairs and Energy within the project ``nxtAIM''. Additionally, the authors gratefully acknowledge the Gauss Centre for Supercomputing e.V. (\url{www.gauss-centre.eu}) for funding this project by providing computing time on the GCS Supercomputer JUWELS Booster at Jülich Supercomputing Centre.